\documentclass[sigconf, nonacm, pdfa]{acmart}
\usepackage{hyperref} 
\usepackage{graphicx}
\usepackage{amsmath}
\usepackage{amsthm}
\usepackage{amsfonts}
\usepackage{subfigure}
\usepackage{subcaption}
\usepackage{balance}
\usepackage[ruled,vlined]{algorithm2e}
\usepackage{algpseudocode}
\usepackage{multirow}
\usepackage{xspace}
\usepackage{enumitem}

\usepackage{pifont}
\newcommand{\one}{\ding{192}} 
\newcommand{\two}{\ding{193}} 

\newcommand{\spara}[1]{\smallskip\noindent{\bf #1}}

\newcommand{\squishlist}{
 \begin{list}{$\bullet$}
  {  \setlength{\itemsep}{0pt}
     \setlength{\parsep}{3pt}
     \setlength{\topsep}{3pt}
     \setlength{\partopsep}{0pt}
     \setlength{\leftmargin}{2em}
     \setlength{\labelwidth}{1.5em}
     \setlength{\labelsep}{0.5em}
} }
\newcommand{\squishlisttight}{
 \begin{list}{$\bullet$}
  { \setlength{\itemsep}{0pt}
    \setlength{\parsep}{0pt}
    \setlength{\topsep}{0pt}
    \setlength{\partopsep}{0pt}
    \setlength{\leftmargin}{2em}
    \setlength{\labelwidth}{1.5em}
    \setlength{\labelsep}{0.5em}
} }

\newcommand{\squishdesc}{
 \begin{list}{}
  {  \setlength{\itemsep}{0pt}
     \setlength{\parsep}{3pt}
     \setlength{\topsep}{3pt}
     \setlength{\partopsep}{0pt}
     \setlength{\leftmargin}{1em}
     \setlength{\labelwidth}{1.5em}
     \setlength{\labelsep}{0.5em}
} }

\newcommand{\squishend}{
  \end{list}
}

\newcommand{\eat}[1]{}

\newcounter{ccc}

\newcommand\redout{\bgroup\markoverwith
{\textcolor{red}{\rule[.5ex]{2pt}{2pt}}}\ULon}

\usepackage{xcolor}
\usepackage{multirow}
\usepackage{array}
\usepackage[a-2b]{pdfx}

\newcommand\vldbdoi{10.14778/XXX}
\newcommand\vldbpages{XXX - XXX}

\newcommand\vldbvolume{XX}
\newcommand\vldbissue{XX}
\newcommand\vldbyear{2026}

\newcommand\vldbauthors{\authors}
\newcommand\vldbtitle{\shorttitle}

\newcommand\vldbavailabilityurl{https://github.com/XXXXXX}

\newcommand\vldbpagestyle{empty}

\begin{document}

\title[\textsf{VikingMem}: A Memory Base Management System for Stateful LLM-based Applications]
{\texorpdfstring{\textsf{VikingMem}: A Memory Base Management System for Stateful LLM-based Applications}
{VikingMem: A Memory Base Management System for Stateful LLM-based Applications}}

\author{Jiajie Fu}
\authornote{Work done while working with ByteDance.}
\authornote{Both authors contributed equally to this research.}
\affiliation{%
  \institution{Zhejiang University}
  \country{China}
}
\email{jiajiefu@zju.edu.cn}

\author{Junwen Chen}
\authornotemark[2]
\affiliation{
  \institution{ByteDance}
  \country{China}
}
\email{chenjunwen@bytedance.com}

\author{Mengzhao Wang}
\authornotemark[1]
\affiliation{
  \institution{Zhejiang University}
  \country{China}
}
\email{wmzssy@zju.edu.cn}

\author{Aoxiang He}
\affiliation{
  \institution{ByteDance}
  \country{China}
}
\email{heaoxiang@bytedance.com}

\author{Maojia Sheng}
\affiliation{
  \institution{ByteDance}
  \country{China}
}
\email{shengmaojia@bytedance.com}

\author{Xiangyu Ke}
\affiliation{%
  \institution{Zhejiang University}
  \country{China}
}
\email{xiangyu.ke@zju.edu.cn}

\author{Yifan Zhu}
\affiliation{%
  \institution{Zhejiang University}
  \country{China}
}
\email{xtf_z@zju.edu.cn}

\author{Yunjun Gao}
\affiliation{%
  \institution{Zhejiang University}
  \country{China}
}
\email{gaoyj@zju.edu.cn}

\begin{abstract}

Large Language Models have revolutionized interactive applications, however, their finite context windows pose a critical data management challenge for maintaining stateful, long-term interactions. 
Existing memory approaches, however, often rely on simplistic extraction that leads to incomplete memories or use rigid, single-use memory extract prompts tailored for one use case like Chatbots. Consequently, they lack generalizability and perform poorly on diverse downstream tasks. 
To bridge this gap, we introduce the Memory Base, a novel data management paradigm to {\em manage the persistent state of long-term interactions}. 
It is characterized by three core principles: {\em selective extraction} of high-value memories from raw information streams; inherent {\em statefulness and evolution}, where memory content is progressively summarized, corrected, and temporally weighted to prioritize recent interactions; and a {\em generalizable abstraction} paradigm designed for robust transferability across diverse applications. 

Building on this foundation, we present {\sf VikingMem}, an end-to-end Memory Base Management System implemented on the VikingDB vector engine. {\sf VikingMem} materializes this paradigm through interconnected event and entity abstractions. It features event-centric memory extraction to selectively handle complex information streams, while entities are dynamically updated by events to achieve stateful evolution. Using temporal compression via a topic-wise timeline and time-weighted recall, the system can progressively produce high-level summary memories, prioritize recent items, while compressing and fading older ones.
Retrieval is further optimized through multi-vector-based reranking. Extensive evaluations on long-term memory benchmarks demonstrate that {\sf VikingMem} achieves state-of-the-art performance, outperforming baselines by up to 30\% in memory retrieval effectiveness while maintaining low latency essential for interactive applications.

\end{abstract}

\maketitle


\pagestyle{\vldbpagestyle}
\begingroup\small\noindent\raggedright\textbf{PVLDB Reference Format:}\\
\vldbauthors. \vldbtitle. PVLDB, \vldbvolume(\vldbissue): \vldbpages, \vldbyear.\\
\href{https://doi.org/\vldbdoi}{doi:\vldbdoi}
\vspace{0.01in}
\endgroup
\begingroup
\renewcommand\thefootnote{}\footnote{\noindent
This work is licensed under the Creative Commons BY-NC-ND 4.0 International License. Visit \url{https://creativecommons.org/licenses/by-nc-nd/4.0/} to view a copy of this license. For any use beyond those covered by this license, obtain permission by emailing \href{mailto:info@vldb.org}{info@vldb.org}. Copyright is held by the owner/author(s). Publication rights licensed to the VLDB Endowment. \\
\raggedright Proceedings of the VLDB Endowment, Vol. \vldbvolume, No. \vldbissue\ %
ISSN 2150-8097. \\
\href{https://doi.org/\vldbdoi}{doi:\vldbdoi} \\
}\addtocounter{footnote}{-1}

\endgroup





\vspace{-4mm}
\section{Introduction}
\label{sec:intro}

Recent advances in large language models (LLMs) have catalyzed a new generation of services — persistent assistants~\cite{brown2020language,touvron2023llama,acharya2023llm}, autonomous agents~\cite{chen2025automatic}, and domain-tailored chat systems~\cite{zhou2025cracking} — that must act coherently across {\em many sessions} and {\em long time horizons}. 
While LLMs' context windows are expanding~\cite{reid2024gemini}
, it remains a finite, costly, and latency-sensitive resource~\cite{santos2024unveiling}. Simply embedding ever-longer histories into every prompt is not only impractical for performance-sensitive services but is architecturally incapable of providing structured, lifecycle-aware state management for consolidation, forgetting, or provenance~\cite{fei2024extending,peng2023yarn}. 
For instance, consider a personalized writing tutor that interacts with a single student over months, retaining past drafts, tracking recurring mistakes, and proactively suggesting targeted exercises. Midway through a long-term engagement, the assistant might repeat questions asked weeks earlier or contradict its own prior corrections. These issues do not primarily stem from limitations in the LLM’s reasoning ability. Instead, they arise because long-term state is often handled in an ad hoc and fragmented manner at the application layer. While an LLM’s context window is transient and bounded, persistent state is frequently spread across logs, shallow caches, and task-specific heuristics rather than maintained through a unified memory mechanism. This fragmented design can lead to broken continuity, stale or conflicting memory, reduced user trust, and substantial developer overhead for custom reconciliation logic~\cite{huang2025survey,li2024snapkv}.
Consequently, applications are {\em shifting from short, stateless queries toward long-lived, stateful interaction patterns}. This shift necessitates a service-grade memory substrate. Here, we draw inspiration from cognitive science, defining ``memory'' not as volatile in-memory storage, but as the persistent, evolving record of experiences that shapes future behavior. Such a substrate must be architected to persistently capture interaction traces, manage their lifecycle through consolidation and updates, enable context-aware retrieval with efficient reranking, and adhere to privacy and audit policies~\cite{jones2024designing,packer2023memgpt}.

This shift necessitates a dedicated memory management substrate, yet existing solutions fail to meet the scale and latency demands of production environments. Early industrial attempts often relied on coarse-grained strategies—such as raw session storage~\cite{li2025hello} or naive topic segmentation~\cite{pan2025secom}—which, while simple, result in bloated storage costs and noise-diluted retrieval. Conversely, recent systems attempting to structure memory via Knowledge Graphs (KGs)—such as ZEP~\cite{rasmussen2025zep} and Mem0's graph integrations~\cite{chhikara2025mem0}—introduce prohibitive operational overhead. For instance, the dynamic KG construction employed by these frameworks incurs significant build-time latency, making them unusable under the high-throughput write loads typical of large-scale services\footnote{In our production environment, individual tenants can generate over 1 billion tokens of memory data daily.}. Furthermore, some systems like Memobase~\cite{Memobase25} are rigidly engineered for narrow verticals—primarily conversational user profiles. This specialization creates a critical adaptability gap: a schema optimized for chat history cannot accommodate the procedural SOPs required by autonomous agents\cite{lyu2024llm,chu2025llm}. Consequently, the industry lacks a unified, configurable memory infrastructure capable of serving these diverse scenarios with a single underlying paradigm.

To address these challenges, we present \textsf{VikingMem}, a unified memory management system architected upon three core design principles derived from our production experience (detailed in \S~\ref{subsec:principles}). First, we argue that effective memory requires \textit{selective information extraction}. Unlike static archives, \textsf{VikingMem} employs fine-grained segmentation to transform low-density raw streams into high-value, discrete signals, maximizing the utility of finite context windows while minimizing token costs. Second, we design the system to possess \textit{inherent statefulness and evolution}. Rather than a static archive, \textsf{VikingMem} treats memory as a living lifecycle where discrete events are managed via temporal weighting to prioritize recent context, while simultaneously driving the progressive consolidation and continuous refinement of persistent entities (e.g., user profiles~\cite{lyu2024llm} or agent SOPs~\cite{chu2025llm}). This dual mechanism ensures that both episodic history and evolving state remain coherent and up-to-date. Finally, to support diverse industrial scenarios—ranging from personalized search and education to complex autonomous agents~\cite{chen2023enterprises,jones2024designing}—we implement a \textit{generalizable design paradigm}. By abstracting memory into a flexible Event-Entity model, \textsf{VikingMem} decouples storage logic from specific business rules, allowing a single infrastructure to serve disparate downstream tasks with minimal reconfiguration.

To this end, we designed and implemented \textsf{VikingMem}, a cloud-native Memory Base Management System built atop the \textsf{VikingDB} vector engine\footnote{\url{https://www.byteplus.com/en/product/vectordatabase}} (a cloud-native vector database used in ByteDance). Architecturally, \textsf{VikingMem} features an intelligent segmentation engine that leverages a two phase LLM-based strategy for precise noise reduction and semantic partitioning (\S~\ref{subsec:memory_ext}), a unified operator library supporting diverse update logic for dynamic entity evolution (\S~\ref{subsec:memory_management}), and a schema-driven extraction pipeline that enables efficient single-pass processing (\S~\ref{subsubsec:components_schemas}), i.e., extracting multiple types of memory from the same input data with a single LLM call. Crucially, this architecture delivers significant efficiency gains, validated through both extensive benchmarks and large-scale production deployment at \textsf{ByteDance} (\S~\ref{sec:exp}). Compared to naive RAG or raw-log approaches, \textsf{VikingMem} reduces storage costs by up to 83.2\% through selective retention and lowers retrieval latency (P95) by 900ms. Furthermore, its single-pass extraction pipeline is highly token-efficient: when extracting $k$ distinct memory types, it reduces token consumption by approximately $(k-1)\times$ compared to multi-pass baselines, where each memory type is extracted with a separate LLM call over the same input. This efficiency enables \textsf{VikingMem} to handle extreme-scale workloads—where a single tenant alone generates over 1 billion tokens daily—with minimal computational overhead. {\sf VikingMem} has been deployed in production and is commercially available. It has supported a broad range of stateful LLM applications like social companionship, efficiency and collaboration tools and personalized education, detailed usage guidelines are in \cite{VikingMem25}. Furthermore, to foster community research and technical knowledge sharing in LLM memory systems, we have open-sourced a subset of the core capabilities of \textsf{VikingMem} via \textsf{OpenViking}\footnote{https://github.com/volcengine/OpenViking}, an open-source Context Database for AI Agents.

\spara{Contribution.}
Our main contributions are summarized below. 
\begin{itemize}[left=0pt]
    \item We formally define the Memory Base, a new architectural framework for persistent state in LLM applications. We distinguish it from prior memory processing methods by establishing three core principles—selective extraction, inherent statefulness and evolution, and a generalizable design paradigm (\S~\ref{sec:intro}). We provide the underlying design schemas of our event-entity model (\S~\ref{sec:system_model_objectives}) to serve as a foundational reference, offering guidance in this fields.

    \item We design and implement {\sf VikingMem}, an end-to-end Memory Base Management System (MBMS) that realizes our proposed framework. We present its detailed design and key technical modules (\S~\ref{sec:details}), such as intelligent memory segmentation strategy, flexible operator-based update mechanism for entity evolution, configurable schema for memory extraction generalizability and other implement details.

    \item We demonstrate the framework's adaptability by detailing its deployment across five distinct industrial scenarios at \textsf{ByteDance}, ranging from autonomous agents to personalized education (\S~\ref{sec:use cases}).

    \item Through extensive experiments on public benchmarks, we demonstrate that {\sf VikingMem} achieves state-of-the-art performance, we demonstrate that \textsf{VikingMem} achieves state-of-the-art performance, outperforming existing baselines by up to 30\% in retrieval accuracy (LLM-Judge Score) maintaining the lowest P95 retrieval latency among all compared methods (\S~\ref{sec:exp}). 
\end{itemize}

\section{System Design Principles and Data Model}
\label{sec:system_model_objectives} 
In this section, we first articulate the industrial observations and core design principles that shape \textsf{VikingMem} (\S~\ref{subsec:principles}). We then formally define the system's foundational data abstractions—the Event-Entity Memory Model (\S~\ref{subsec:event_entity_model})—and finally outline the key design objectives that guide its implementation (\S~\ref{subsec:design_objectives}).

\vspace{-2mm}
\subsection{Industry Observations and Design Principles}
\label{subsec:principles}

Based on our extensive experience serving diverse enterprise clients and commercializing LLM solutions at \textsf{ByteDance}, we identify three critical requirements that a production-grade memory substrate must satisfy. These observations directly motivate the core capabilities of \textsf{VikingMem}.

\vspace{-1mm}
\spara{Observation 1: Low Signal-to-Noise Ratio in Raw Streams.} 
Unlike traditional RAG systems that ingest high-density, pre-processed documents~\cite{gan2025retrieval,karpukhin2020dense}, production memory systems operate on low-density raw streams (e.g., meeting transcripts, debugging logs) where valuable information is sparse and topics are interleaved~\cite{han2024let,wang2025mirix}. Simply truncating or summarizing these streams blindly leads to ``context pollution,'' where LLMs are distracted, or ``over-squashed'', by irrelevant noise~\cite{barbero2024transformers}. 

\noindent \textbf{Principle 1:} To address this, \textsf{VikingMem} is built upon \textit{selective information extraction}. The system must go beyond naive chunking and provide \textit{fine-grained event extraction and segmentation}. 
Specifically, this requires a sophisticated strategy to prune irrelevant noise (e.g., greetings, fillers) and identify the precise start and end indices of semantic segments, reconstituting disjointed information into coherent ``events'' (\S~\ref{subsec:memory_ext}). Furthermore, given the high token cost of raw streams, this extraction must be \textit{computationally efficient}. The system should employ a low-cost technique that minimizes redundant processing when extracting multiple memory types from the same stream, avoiding the excessive token consumption of traditional multi-pass approaches~\cite{chhikara2025mem0}.

\spara{Observation 2: Dynamic Nature of State.} 
Real-world entities — whether a user learning a language or an agent executing a workflow—are constantly changing. Traditional memory approaches that rely on static vector storage (i.e., simple "insert-and-retrieve" RAG pipelines) fail to capture this \textit{evolution}. They merely accumulate history without updating the underlying state, leading to contradictions and stale information.

\noindent\textbf{Principle 2:} Drawing inspiration from cognitive science~\cite{markus1977self,linville1987self}, we design \textsf{VikingMem} to possess \textit{inherent statefulness and evolution}. It treats memory as a living lifecycle where discrete events actively drive the continuous refinement of persistent entities, mirroring human memory consolidation processes~\cite{roediger2006test}. 
To realize this, the system must support \textit{dynamic entity-state tracking} through a unified update mechanism. This involves a library of reusable \textit{operators} (e.g., \texttt{LLM\_MERGE}, \texttt{SUM}, \texttt{TIME\_COMPRESS}) that programmatically link events to entities (\S~\ref{subsec:memory_management}). These operators handle both efficient updates and progressive consolidation, strategically summarizing or forgetting dated information~\cite{wu2025unfolding} to maintain a compact state. Furthermore, to ensure long-term relevance during retrieval, the system applies dynamic weighting mechanisms (including temporal and business-specific weights) to prioritize the most pertinent context (\S~\ref{subsec:memory_management}).

\spara{Observation 3: Fragmentation of Memory Architectures.} 
Different downstream applications demand fundamentally disparate memory structures and contents. For instance, emotional companions focus on capturing user personality and conversational style from dialogue history~\cite{jones2024designing}, whereas autonomous agents require extracting executable experiences from historical trajectories to refine their standard operating procedures (SOPs)~\cite{wang2024agent}. Building siloed solutions for each scenario leads to high maintenance costs and prevents capability sharing.

\noindent\textbf{Principle 3:} To support these diverse scenarios, \textsf{VikingMem} adopts a \textit{generalizable design paradigm}. We implement a flexible and transferable memory schema that decouples the underlying storage logic from specific business rules.
By abstracting memory into a configurable \textit{Event-Entity model} defined via a JSON-like schema (\S~\ref{subsubsec:components_schemas}), a single infrastructure can represent disparate memory types (e.g., declarative preferences, procedural SOPs) and serve downstream tasks with minimal reconfiguration. This schema-driven design is also the key enabler for the single-pass extraction efficiency mentioned in Principle 1.

\subsection{Event-Entity Primitives}
\label{subsec:event_entity_model}

The core of \textsf{VikingMem}'s architecture is built upon two fundamental data abstractions, Events and Entities. This section defines these components and the mechanism that links them.

\vspace{-1.2mm}
\subsubsection{Core Components and Schemas}
\label{subsubsec:components_schemas}

In this part, we will formally define the fundamental components that form the basis of our model: the Event and Entity data abstractions. We will detail their respective schemas, which specify their structure and—critically for the Entity—embed the Operator logic that links the two.

\begin{figure}[!tb] 
    \centering 
    \includegraphics[width=0.47\textwidth]{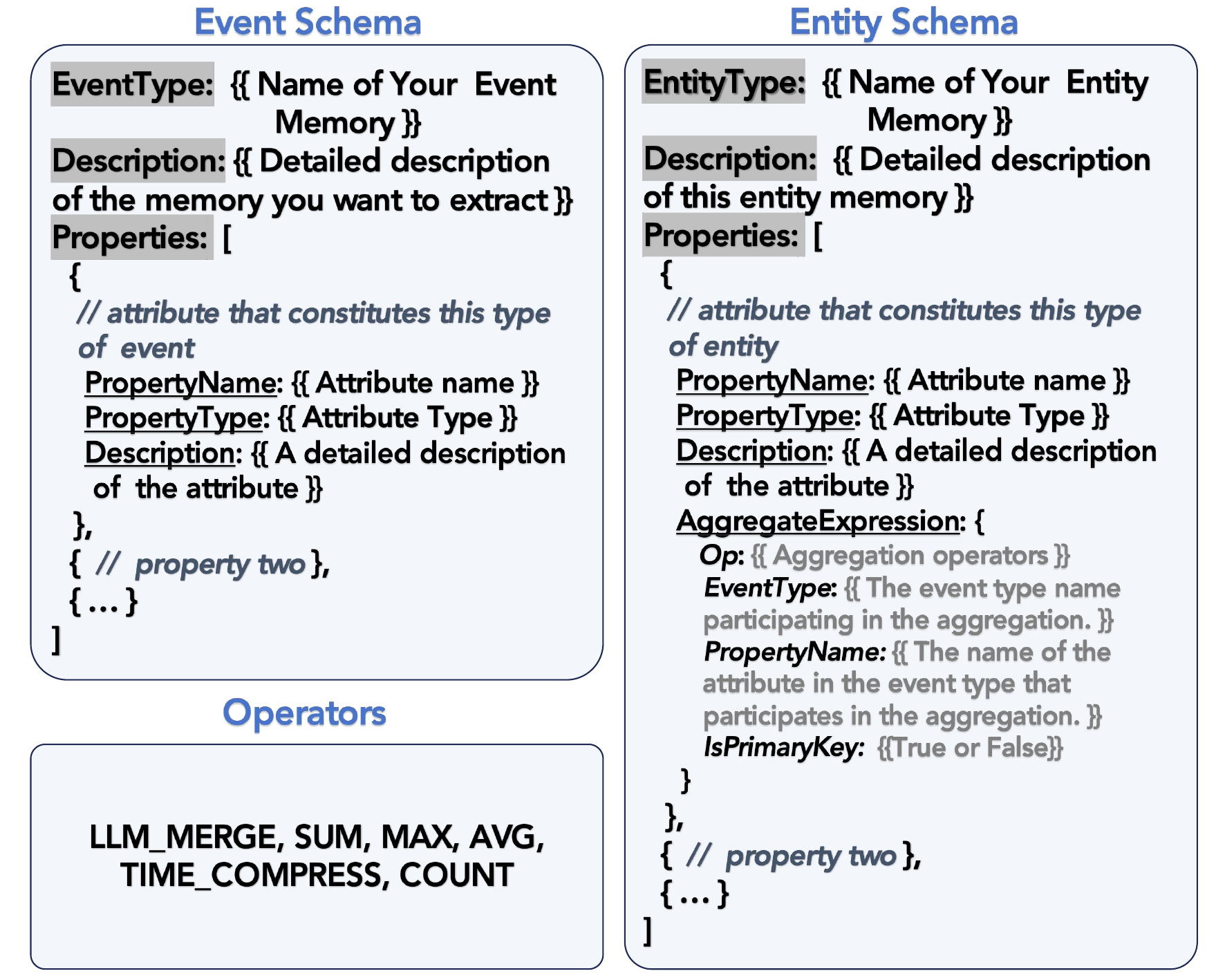} 
    \vspace{-3mm} 
    \caption{The definition schema of event, entity for memory extraction, and the built-in operators in \textsf{VikingMem}} 
    \vspace{-4mm} 
    \label{fig:schema} 
    \vspace{-2mm}
\end{figure}

\spara{Event:}  
An event is the foundational unit of memory, representing a discrete, timestamped, and episodic record.\footnote{Human memory is fundamentally event-based and organized into discrete episodic records \cite{gonneaud2014we}.} The full event instance includes immutable metadata, such as a timestamp; these elements are omitted from Figure \ref{fig:schema} for clarity. A detailed example of the complete event instance can be found in \S~\ref{subsec:memory instance}.
They are the unit of ingestion: compact, meaningful records selectively extracted from raw interaction streams (sessions) that capture a single, salient piece of information.
The Event Definition Schema shown in Figure \ref{fig:schema} serves as a customizable template, providing users with significant flexibility to tailor the extraction process to their application needs by specifying an {\sf EventType}, {\sf Description}, and a set of {\sf Properties}.
Regarding each property, we also allow users to define its own {\sf PropertyName}, {\sf PropertyType}, and a detailed {\sf Description}.
This strict schema both constrains and empowers the memory extraction process: it prevents noisy, unimportant information from entering the memory base and guides LLMs to extract memories in a structured and consistent manner. 
Importantly, an event in {\sf VikingMem} is not a raw recency-based context window, but a selectively extracted, semantically coherent, schema-constrained memory unit derived from potentially interleaved raw sessions.

\spara{Entity:} An entity represents a persistent, evolving state representation (e.g., a comprehensive user profile, an agent's library of Tool Usages). Unlike instantaneous, episodic events, entities integrate and consolidate information over time to form a coherent, long-term memory. The Entity Schema (right part in Figure \ref{fig:schema}) grants similar flexibility, allowing users to define an {\sf EntityType}, {\sf Description}, and its constituent Properties. The core mechanism for this stateful evolution is detailed within each property's {\sf AggregateExpression}. This expression explicitly dictates the rule for property updates by defining the required {\sf EventType} and {\sf PropertyName} from an incoming event, thereby specifying exactly which events trigger the entity's evolution.
Likewise, an entity is not merely a compacted status note, but a persistent state representation incrementally materialized from events through explicit aggregation expressions and operators.

\spara{Operator:} An operator is an {\em application-defined} function that governs {\em how an entity's state is updated by an associated event}. 
Operators abstract the business logic of memory evolution. When a linked event is ingested, the operator specified in the {\sf AggregateExpression}'s {\sf Op} field is executed, using the designated event attribute to update the entity's related attributes. This process completes the memory's update and evolution cycle, ensuring the entity remains a coherent state. As shown in Figure \ref{fig:schema}, \textsf{VikingMem} provides a set of built-in operators (e.g., \texttt{SUM}, \texttt{MAX}, \texttt{AVG}, \texttt{COUNT}, \texttt{LLM\_MERGE}) that can aggregate numeric data, overwrite old values, or perform complex textual merges (the details of the operators can be found in \S \ref{subsec:memory_management}).

\subsubsection{Demystifying the Event-Entity Paradigm}
\label{subsec:demystifying}

The rationale for this event–entity abstraction is its ability to serve as a general substrate for diverse downstream tasks. Traditional memory architectures are, in essence, built around a single notion of short-lived, event-level (or “episode”) memory~\cite{wang2025mirix,chhikara2025mem0}. The additional “memory types” they introduce are usually not new abstractions: they either (i) change the prompts used for event extraction, or (ii) hard-code task logic into prompts, so that each business workflow encodes its own event selection, summarization, and update rules directly in LLM prompts. As a result, {\em every new scenario requires another layer of prompt engineering}, while the memory system itself does not provide a stable, reusable interface. In contrast, {\sf VikingMem} treats the event store as a generic log and the entity store as a family of materialized views over that log. The relationship between events and entities is expressed as a parametric query, rather than a hand-crafted prompt, as illustrated below:

\vspace{-3mm}
\begin{align*}
\texttt{entity}
:=\;&
\operatorname{SELECT}\ \mathsf{OP}\bigl(\texttt{event.content}\bigr) \operatorname{FROM}\ \texttt{Events} \\
&\operatorname{WHERE}\ \mathrm{filters}\bigl(\texttt{event}\bigr) \operatorname{GROUP\ BY}\ \mathrm{keys}\bigl(\texttt{event}\bigr).
\end{align*}
\vspace{-3mm}

Here, \textsf{keys} defines how events are grouped into entities (e.g., at the per-user level, per user–assistant pair, or per topic), \textsf{filters} constrains which events are eligible (e.g., a time window), and $\mathsf{OP}$ is chosen from a fixed operator library (e.g., \texttt{LLM\_MERGE}, \texttt{TIME\_COMPRESS}, \texttt{AVG}, \texttt{SUM}). Application-specific prompts appear only at the edges of this pipeline—within event extraction and, where necessary, inside LLM-related operators—while the overall transformation pattern remains unchanged.

This design yields both abstraction and flexibility. By changing only \textsf{keys}, \textsf{filters}, and the selected operator, the same event log can materialize user profiles, SOPs, agent working memories for tools, search–recommendation–QA traces, or subject-level learning progress records, without modifying the core memory engine. Moreover, the same operators are reused across scenarios: an \texttt{LLM\_MERGE} operator over “user–day” groups produces periodic entity snapshots, while the same \texttt{LLM\_MERGE} operator over “user” groups realizes incremental, lifelong entity refinement. In this sense, {\sf VikingMem} elevates memory design from prompt-by-prompt engineering to a small, reusable algebra over events and entities, making the system inherently reusable, programmable, and compatible with diverse downstream applications.

\begin{figure*}[t] 
    \centering 
    \includegraphics[width=0.9\textwidth]{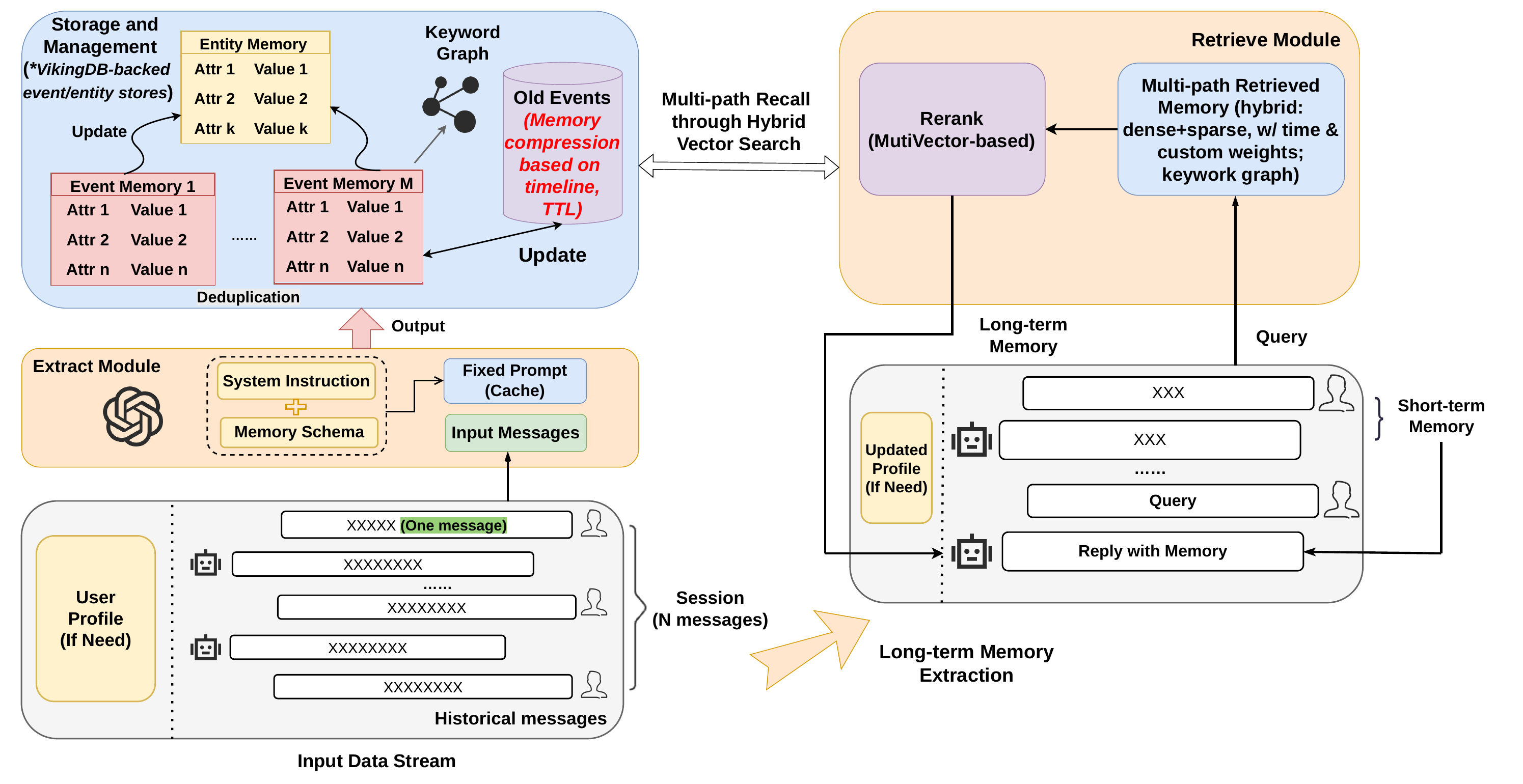} 
    \vspace{-3mm} 
    \caption{ The pipeline of \textsf{VikingMem}} 
    \label{fig:echo_pipeline} 
    \vspace{-2mm}
\end{figure*}

\vspace{-1mm}
\subsection{Design Objectives}
\label{subsec:design_objectives}

The design of \textsf{VikingMem} is driven by the central challenge of balancing the competing objectives of maximizing the utility of memory for downstream LLM tasks and minimizing the operational costs of the memory lifecycle. We characterize this challenge via two primary objectives:

\textbf{\one\ Memory Effectiveness:} This refers to the quality and utility of the memory context provided to the LLM $\mathcal{L}$. The system's goal is to process an input query $Q$ and retrieve/construct a memory context $M$ such that the performance of the LLM $\mathcal{L}(Q, M)$ (e.g., response accuracy, continuity, and relevance) is maximized. An effective memory $M$ bridges the gap between the LLM's bounded, transient context and the ideal, "all-knowing" state required for a perfect response.

\textbf{\two\ Operational Efficiency:} This encompasses the full-stack costs associated with the memory lifecycle. A service-grade system must aggressively minimize: (1) Extraction Cost: The computational overhead of processing low-density raw data streams to selectively identify and store salient events. (2) Retrieval Latency: The time-to-first-token impact of fetching, reranking, and consolidating relevant memories into a prompt, which is critical for performance-sensitive, real-time applications.

An effective memory system must balance these competing objectives, providing high-value, relevant memories without introducing prohibitive latency or cost.

\vspace{-1.8mm}
\section{System Designs}
\label{sec:details}
In this section, we provide a detailed explanation of the {\sf VikingMem} system design and its core components. As illustrated in Figure~\ref{fig:echo_pipeline}, the system is architected around a central pipeline comprising three key stages: Memory Extraction (\S~\ref{subsec:memory_ext}), Memory Management (\S~\ref{subsec:memory_management}), and Memory Retrieval (\S~\ref{subsec:memory_retrieve}). {\sf VikingMem} is implemented as a memory-management layer over {\sf VikingDB}, which serves as the underlying storage and retrieval substrate. Concretely, event memories are persisted as searchable records with text, metadata, and hybrid retrieval features; entity memories are maintained as separately updated state records; and the keyword graph is materialized as an auxiliary association index. In this design, {\sf VikingDB} is responsible for persistent storage while {\sf VikingMem} implements the memory-specific logic above it, such as schema-driven extraction, deduplication and operator-based event-to-entity updates. The Memory Extraction module processes raw data to capture salient information; the Memory Management module handles the memory lifecycle, including event-to-entity updates via operators and progressive consolidation; and the Memory Retrieval module selects relevant memories to provide context for a stateful response. In the following, we detail the core technical designs, optimizations, and motivations behind each component.

\vspace{-4mm}
\subsection{Memory Extract}
\label{subsec:memory_ext}

As illustrated in Figure~\ref{fig:echo_pipeline}, the memory extraction module receives a session (e.g., a batch of $N$ messages) from the input data stream and processes it as input messages to perform memory extraction. This process is activated once a logical batch of data has been accumulated (i.e., a completed session). While this batching can be triggered dynamically, we find that a well-defined threshold ($\ge 20$ messages) for data accumulation often yields the most consistent and high-quality memories . This design is also crucial for balancing short-term and long-term memory. We define short-term memory as the most recent data within the user's active session, which is retained for immediate contextual relevance. Meanwhile, this session is then converted into persistent long-term memories. This conversion prevents the LLM's active context from becoming overloaded—a condition that can lead to hallucinations and heightened computational complexity~\cite{huang2025survey,liu2025comprehensive}. By systematically converting accumulated data into persistent long-term memories, our approach preserves contextual richness while enabling real-time updates. 

Next, we will first detail our one-pass memory extraction paradigm. We elucidate how we leverage the structured event and entity schemas to instruct a large language model to complete the entire memory workflow within a single inference pass. This entire mechanism is built upon the LLM's in-context learning (ICL) ability. Afterwards, we describe the prefix-cache optimizations we designed for the memory extraction prompt, i.e., reusing the cached representations of the fixed prompt prefix (such as the system instruction and memory schema) across requests~\cite{pan2025kvflow}. It substantially reduces token consumption during the process. Finally, we explain how we leverage event schemas and sophisticated memory segmentation technique to derive more comprehensive and meaningful memory information from low-information-density input data stream.

\begin{figure}[t]
    \centering
    \includegraphics[width=0.45\textwidth]{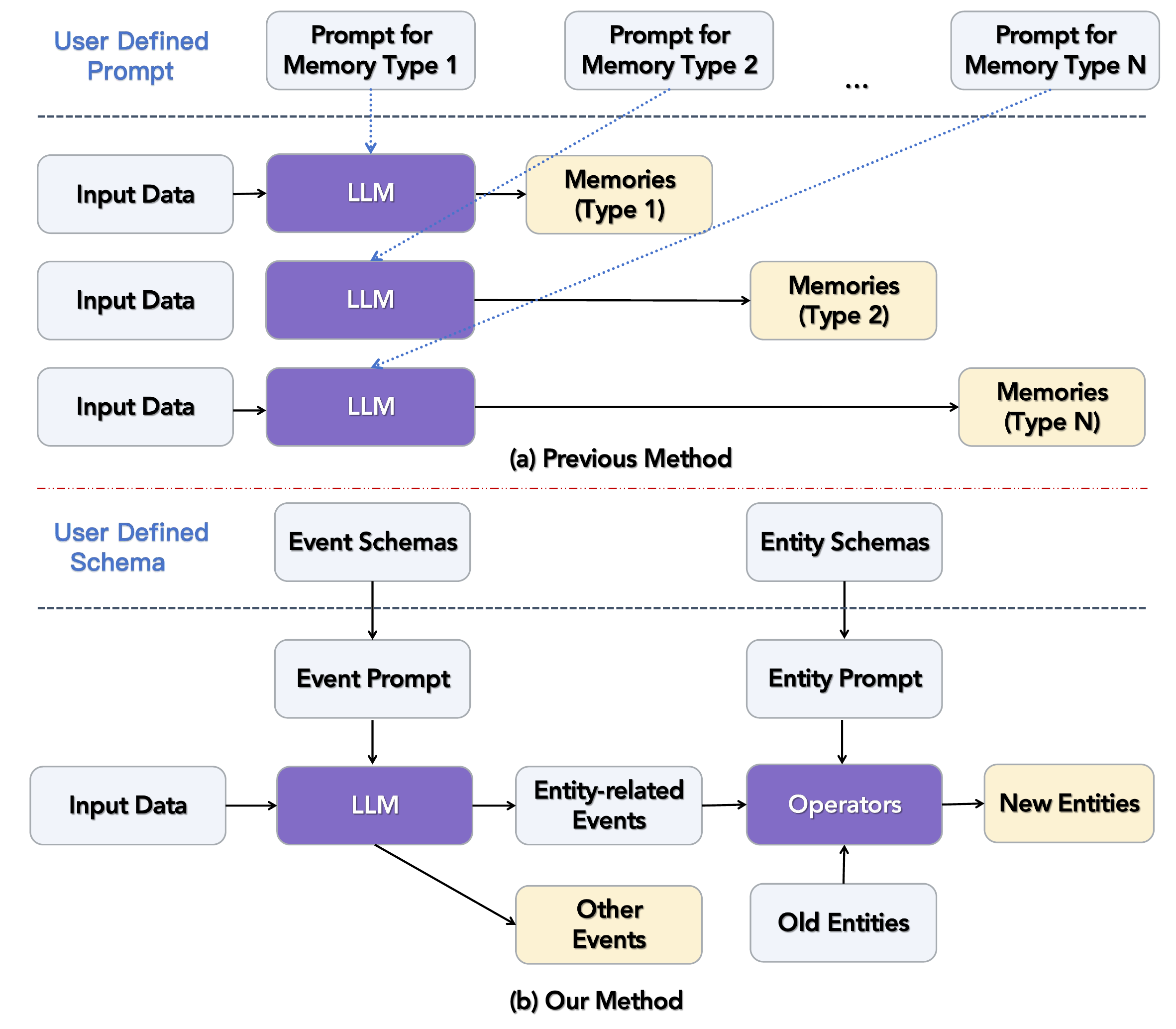}
    \vspace{-3.3mm}
    \caption{ The extract paradigm in \textsf{VikingMem}}
    \label{fig:memory_extract}
    \vspace{-3.4mm}
\end{figure}

\vspace{-2mm}
\spara{One-pass Memory Extraction.} Traditional memory extraction paradigms, as illustrated in Figure~\ref{fig:memory_extract}, suffer from significant inefficiencies. Prior works, such as \cite{Memobase25,wang2025mirix}, typically define each memory type using a distinct, hard-coded prompt. This not only demands extensive prompt engineering and yields inconsistent results, but more critically, it necessitates a multi-pass approach: to extract $N$ memory types, the same raw input data must be fed to the LLM $N$ times. This repeated processing leads to prohibitive token consumption and computational costs, far exceeding practical industrial limits. In contrast, \textsf{VikingMem} implements a \textit{one-pass memory extraction paradigm}. Instead of multiple unstructured prompts, we leverage our configurable event and entity schemas (\S~\ref{subsubsec:components_schemas}). These schemas are programmatically compiled into a single, comprehensive prompt. This allows the LLM to process the raw input stream just once and, in that single pass, simultaneously extract all defined memory types—both new events and the precise content for entity updates. This schema-driven, one-pass approach achieves high-quality, structured extraction while drastically reducing computational costs, making complex memory operations efficient and economically viable (demonstrated in \S~\ref{subsec:exp_extraction}).

\begin{figure}[t]
    \centering
    \includegraphics[width=0.45\textwidth]{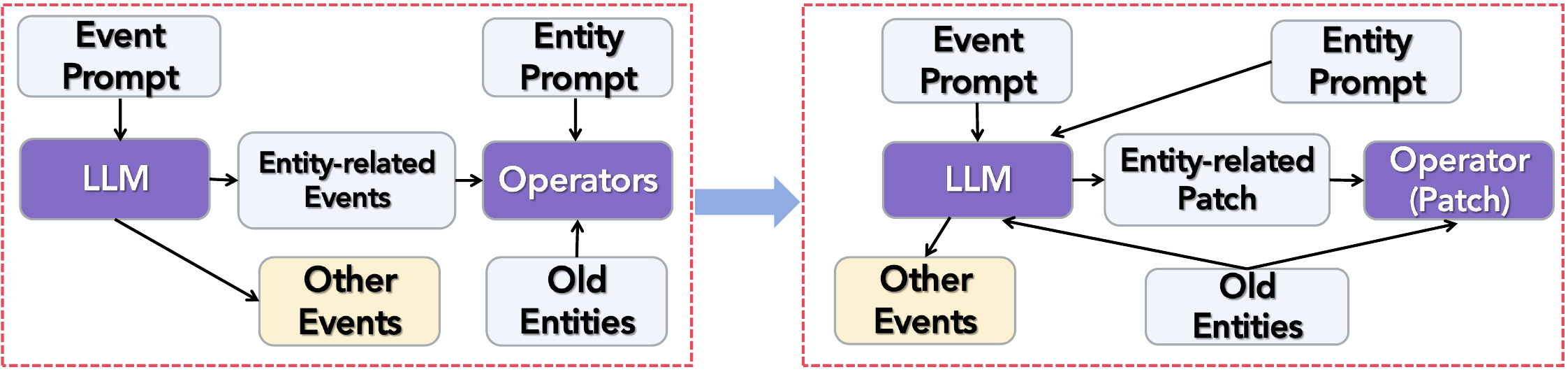}
    \vspace{-2mm}
    \caption{ Faster Entity Update w/o LLM}
    \label{fig:prompt}
    \vspace{-5mm}
\end{figure}

\vspace{-1mm}
\spara{Faster Entity Update.}
Although \textsf{VikingMem} can extract all memory types in a single LLM pass, updating \textbf{entity} memories may still introduce an extra LLM call when an entity field is of \emph{string} type: we typically feed the extracted \emph{entity-related events} together with the old entity back to the LLM to synthesize the updated entity. This is acceptable for offline processing, but in latency-sensitive scenarios (e.g., real-time, in-dialogue knowledge/attribute summarization where users only care about entity memory), a second LLM call significantly increases both response time and cost. To enable online, low-cost updates, we propose an \underline{E}ntity \underline{U}pdate \underline{A}lgorithm (\textsf{EUA}, Algorithm \ref{alg:eua}) inspired by Roocode~\cite{Roocode}, which updates entities \emph{without any additional LLM}. Specifically, for each entity field defined in the schema, the extractor outputs a compact \emph{patch} in the form ``\texttt{<<<< SEARCH ... ==== ... >>>> REPLACE}''; \textsf{EUA} then applies this patch to the corresponding old field by locating the best-matching span using \textbf{edit-distance-based approximate search}~\cite{deng2013top} (robust to minor LLM inaccuracies that break exact string matching) and replacing it with the \texttt{REPLACE} content. This turns entity maintenance into a lightweight deterministic post-processing step: the one-pass extraction produces field-wise patches, and EUA applies them to obtain the new entity instantly, achieving near-perfect update reliability with drastically reduced latency and token cost. Importantly, this patching process does not require enumerating all stored entities. For each update, VikingMem first retrieves only a small candidate set of relevant existing entities (top-5 in our deployment) using ANN search over entity representations, and includes only these candidates in the prompt for patch generation. This keeps the prompt length bounded while preserving sufficient context for accurate updates.

\setlength{\textfloatsep}{0pt}
\begin{algorithm}[t]\LinesNumbered
\small
\caption{EUA: Patch-based Entity Update w/o LLM}
\label{alg:eua}

\KwIn{Entity schema $\mathcal{S}$; old entity $E^{old}$; field-wise patches $\{p_f\}$}
\KwOut{Updated entity $E^{new}$}

$E^{new}\leftarrow E^{old}$\;
\ForEach{field $f$ in $\mathcal{S}$}{
  $(s,r)\leftarrow \textsc{ParsePatch}(p_f)$ \tcp*{$s$=\texttt{SEARCH}, $r$=\texttt{REPLACE}}
  \If{$s=\emptyset$}{\textbf{continue}}
  $(i,j)\leftarrow \textsc{BestApproxSpan}(E^{old}[f], s)$ \tcp*{min edit distance}
  $E^{new}[f]\leftarrow E^{old}[f]_{[0:i]} \,\Vert\, r \,\Vert\, E^{old}[f]_{[j:]}$\;
}
\Return{$E^{new}$}\;
\end{algorithm}
\setlength{\textfloatsep}{12pt plus 2pt minus 2pt}



\vspace{-1mm}
\spara{Intelligent Memory Segmentation for Event-intertwined Sessions.}
In memory extraction, input data streams (particularly conversational ones) often exhibit low information density and interleaved topic discussions (e.g., shifting from topic A to B, then back to A)~\cite{han2024let}. Traditional methods are suboptimal: message-level storage creates fragmented memories, session-level storage introduces excessive noise, and sequential topic-level approaches fail to merge non-contiguous segments~\cite{ye2024boosting,pan2025secom}.
To illustrate, consider the example
where a conversation on recursion (topic A) is interrupted by a game algorithm discussion (topic B). For a query about recursion, fragmented retrieval (A1, B, A2) would yield incomplete answers, while session-level consolidation would introduce noise from B, leading to factually inaccurate outputs.
In \textsf{VikingMem}, we introduce a novel Intelligent Memory Segmentation Strategy to address these challenges. This strategy leverages an LLM to execute a two-phase process. The first phase, semantic saliency filtering, identifies and isolates only the most meaningful segments from the raw dialogue. This effectively prunes conversational filler, such as initial greetings, that are not worth storing—a significant advantage over traditional RAG chunking strategies. This ensures only high-value information is processed, reducing both noise and storage overhead. In the second phase, event-centric partitioning, the LLM determines the precise start and end positions for each coherent topic, outputting them as a list of tuples, a format that is not only precise but also highly token-efficient.
This coordinate-based mapping allows for the consolidation of semantically related but non-contiguous dialogue segments. This mechanism enables the formation of more complete and cohesive memories by consolidating scattered content from the same topic. For the illustrated query, our method accurately merges A1 and A2 into a unified memory while excluding B's noise, thereby producing a fully correct and comprehensive response that \textit{encompasses the recursion definition, base case warnings, merge sort application, and efficiency analysis without distortions.}


\vspace{-2.5mm}
\subsection{Memory Management}
\label{subsec:memory_management}

The memory management module, as illustrated in Figure~\ref{fig:echo_pipeline}, receives extracted memories and processes them for efficient storage and updating. This module encompasses three core functionalities: (1) event memory management, (2) entity memory management, and (3) keyword graph updates. 
We next present event handling, entity updates, time-based compression, and keyword graph construction.


\vspace{-1.5mm}
\spara{Event Memory Management.} Upon extracting new memories, the event update process begins with deduplication to ensure efficiency and accuracy~\cite{fu2025context}. Specifically, within the same session, highly similar memories are identified and merged to eliminate redundancy by using LLM-based resolution~\cite{rasmussen2025zep}. 

\vspace{-1.5mm}
\spara{Entity Memory Update by Novel Operators.} As outlined in Section \S~\ref{subsubsec:components_schemas}, entity memories are aggregated from related event memories via a suite of operators implemented in \textsf{VikingMem}, which we abstracted from practical applications. These operators, including \texttt{SUM}, \texttt{COUNT}, \texttt{MAX}, \texttt{AVG}, \texttt{LLM\_MERGE}, and \texttt{TIME\_COMPRESS}, are broadly categorized into statistical and LLM-based. 

The statistical operators (\texttt{SUM}, \texttt{COUNT}, \texttt{AVG}, \texttt{MAX}) operate without invoking large language models, focusing on straightforward computations—such as calculating a user's accuracy rate in educational exercises or tracking behavior patterns. This approach addresses the known limitations of LLMs in arithmetic tasks~\cite{zhang2024counting}, thereby reducing token usage while ensuring greater precision. In contrast, the LLM-based operators handle complex synthesis. The \texttt{LLM\_MERGE} operator leverages an LLM to perform incremental updates, handling tasks like content deduplication and resolving conflicts between new and existing information. 
The \texttt{TIME\_COMPRESS} operator provides a mechanism for long-term memory lifecycle management, implementing our ``progressive consolidation'' design. Instead of treating memories as a flat log, it groups related events into topic-centric timelines and orders them chronologically, which enables a ``lazy merge'' strategy. Recent memories are retained with high fidelity, whereas older events in an inactive timeline are synthesized only after they age into a predefined temporal window (e.g., forming a weekly or monthly summary), emulating human memory mechanisms~\cite{mijalkov2025computational}. After such synthesis, the system stores the resulting higher-level summary and assigns a Time-To-Live (TTL) to the underlying fine-grained events. Once the summary has preserved their salient information, expired low-level events can be safely pruned unless subsequent recalls reinforce that timeline. In this way, \texttt{TIME\_COMPRESS} balances contextual richness with storage and retrieval efficiency.
To enhance the decision-making and reliability for both \texttt{LLM\_MERGE} and \texttt{TIME\_COMPRESS}, we incorporate techniques such as chain-of-thought~\cite{wei2022chain} and one-shot~\cite{brown2020language} for more nuanced memory revisions.

\spara{Keyword Graph.} \textsf{VikingMem}'s default memory retrieval employs a dense-sparse hybrid approach~\cite{zhang2024efficient}, yet it inadequately addresses queries such as ``Do you remember my nickname?'' This limitation arises as 
the semantic similarity between the direct query and the indirect memory is very low. Drawing from~\cite{huang25kdd}, we construct a keyword graph where each keyword's embedding is averaged from the embeddings of memory segments containing it, better capturing semantic features, with keywords connected to associated memory.

\vspace{-2mm}
\subsection{Memory Retrieve}
\label{subsec:memory_retrieve}
Within the workflow outlined in Figure~\ref{fig:echo_pipeline}, the memory retrieval module takes user queries as input and selects appropriate memories (with time-weighting) from the organized memory repository mentioned above.


\vspace{-2mm}
\spara{Multi-path Recall with Time-Decay and Business Weights.} In \textsf{VikingMem}, memory recall is divided into two paths to enhance retrieval effectiveness. The primary path performs hybrid retrieval directly over the memory store by combining dense semantic matching and sparse lexical matching for robust results~\cite{zhang2024efficient}. Beyond the original retrieval score, this path incorporates two explicit priors: \textit{recency} via a time-decay score and \textit{business importance} via a business-specific score. Specifically, for each candidate memory we compute a final score $S_{\text{final}}$ as a weighted combination of the normalized original retrieval score $S_{\text{origin}}$, the temporal score $S_{\text{time}}$, and the business score $S_{\text{busi}}$ (all normalized to $[0,1]$): $S_{\text{final}}=(1-w_{\text{time}}-w_{\text{busi}})\cdot S_{\text{origin}}+w_{\text{time}}\cdot S_{\text{time}}+w_{\text{busi}}\cdot S_{\text{busi}}$, where $w_{\text{time}},w_{\text{busi}}\in[0,1]$ are configurable and satisfy $w_{\text{time}}+w_{\text{busi}}\le 1$. The temporal score $S_{\text{time}}$ is computed from the memory age with a user-configurable freshness tolerance window (e.g., 1 day) where recent memories receive $S_{\text{time}}=1$; for memories older than this window, $S_{\text{time}}$ decays exponentially following a fast-then-slow curve. For business importance, we support two complementary definitions of $S_{\text{busi}}$: (i) \textit{type-level weighting}, where users assign different importance weights to different event-memory types to prioritize critical categories during retrieval; and (ii) \textit{instance-level weighting}, where a dedicated weight field is included in the memory schema and extracted together with the event to represent the importance of a specific memory instance within the same type, which is then used to promote high-value instances at recall time. Meanwhile, the auxiliary path leverages a keyword graph to retrieve complementary memories, addressing scenarios where standard hybrid search may fall short~\cite{huang25kdd}. Through empirical evaluations, we observed that simply merging results from both paths yields inferior performance compared to independently ranking each path's outputs, assigning distinct quotas (with a larger allocation to the primary path), and then merging the reranked selections, which leads to a more balanced and higher-quality final set of retrieved memories.

\vspace{-1mm}
\spara{Multi-vector Rerank.} In memory retrieval scenarios, end-to-end latency is subject to stringent requirements, with reranking operations typically needing to achieve p99 latencies in the order of hundreds of milliseconds. Certain cross-encoder-based reranking algorithms, such as mGTE~\cite{zhang2024mgte}, often result in p99 latencies in the order of seconds, especially when the candidate memory set is relatively large, rendering them unsuitable for real-time applications. To address this, we adopt a reranking strategy inspired by ColBERT~\cite{khattab2020colbert}, which enables efficient and effective memory search through contextualized late interaction. Specifically, we precompute and store ColBERT vectors for memories during the extraction step, applying a series of compression techniques including quantization~\cite{gao2024rabitq,kuzmin2022fp8}, and token-merge operations~\cite{ma2025storage}. These optimizations ensure that storage overhead remains comparable to that of dense vectors, thereby facilitating efficient reranking with minimal additional storage costs.
\vspace{-1.5mm}
\section{Applications And Example Use Cases}
\label{sec:use cases}

\subsection{Downstream Application Scenarios}

As introduced in \S~\ref{sec:intro}, {\sf VikingMem} is designed to support a broad range of stateful LLM applications. We summarize our primary deployment domains into five representative categories. These capabilities can be experienced via our commercial services~\cite{VikingMem25}.

\vspace{-0.5mm}
\spara{Social \& Companionship.} Personalized chatbots~\cite{jones2024designing,wu2024longmemeval} use {\sf VikingMem} to capture salient episodic memories and user preferences, enabling more contextual and empathetic responses.

\vspace{-0.5mm}
\spara{Search \& Recommendation.} Search and recommendation systems~\cite{lyu2024llm,acharya2023llm} use {\sf VikingMem} to consolidate cross-domain activity traces into evolving user profiles for long-term personalization.

\vspace{-0.5mm}
\spara{Efficiency \& Collaboration.} Meeting assistants and note-taking tools~\cite{chen2023enterprises} use {\sf VikingMem} to extract key points and actionable to-dos from unstructured streams and consolidate them into periodic reports for better efficiency.

\vspace{-0.5mm}
\spara{Education.} Personalized learning systems~\cite{swacha2025retrieval,chu2025llm} use {\sf VikingMem} to track fine-grained mastery of knowledge points and support adaptive exercises.

\vspace{-0.5mm}
\spara{Agent Memory.} Autonomous agents use {\sf VikingMem} to capture execution trajectories and tool usage logs~\cite{wu2025meta,xiao2025toolmem}, progressively refining reusable tool experience and standard operating procedures (SOPs) for future tasks.


\vspace{-2.5mm}
\subsection{Example Event-Entity Instance}
\label{subsec:memory instance}

To provide a concrete understanding of how our abstract event-entity model is applied in practice, we focus on one representative application scenario in greater detail. Specifically, we use {\sf Agent Memory} to illustrate how raw interaction streams are transformed into structured event instances and further consolidated into persistent entities. More comprehensive use case examples are provided in~\cite{usecases}.

\begin{figure}[!tb] 
    \centering 
    \includegraphics[width=0.42\textwidth]{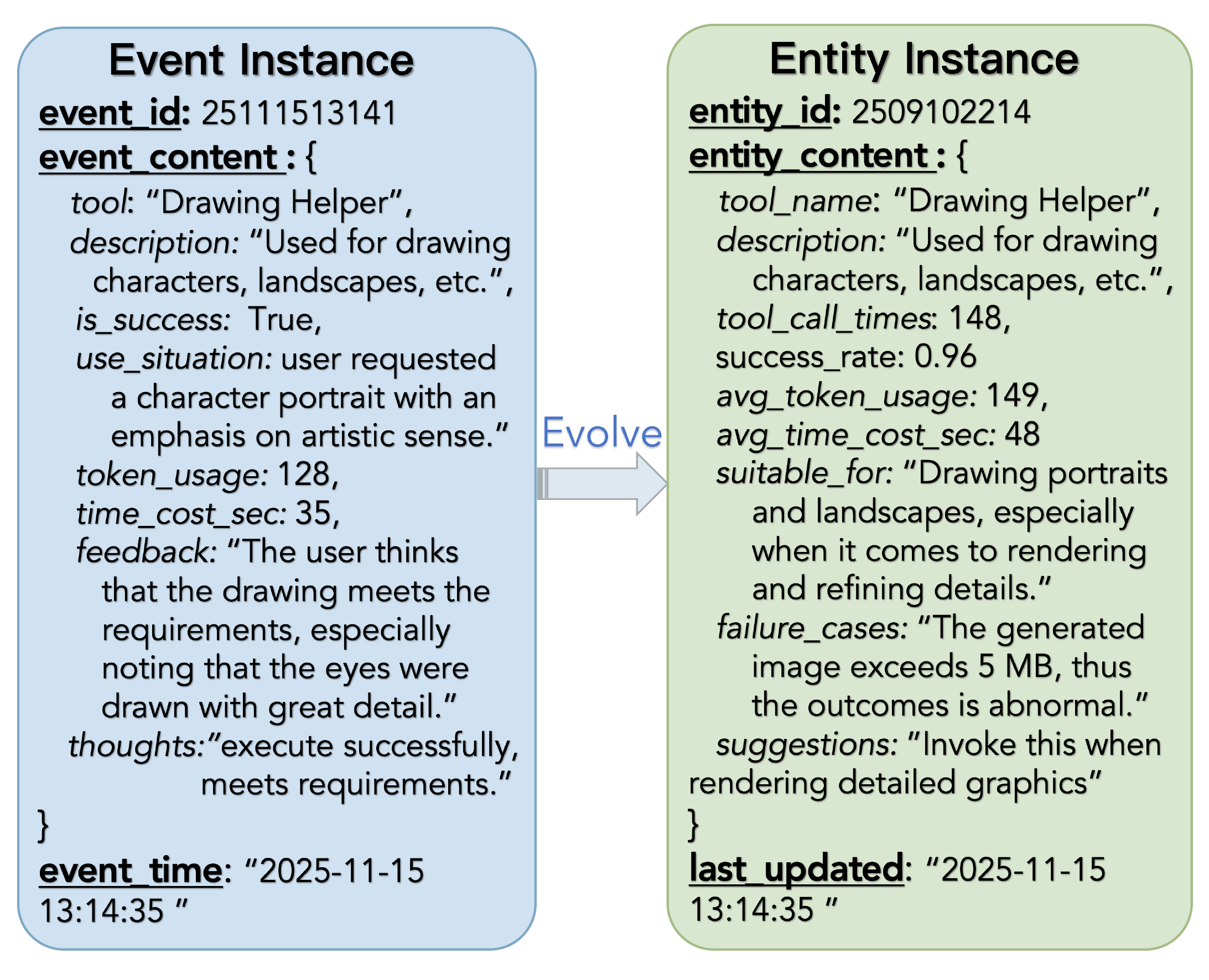} 
    \vspace{-3mm} 
    \caption{Use case for Agent Memory} 
    \vspace{-3mm} 
    \label{fig:agent_case} 
    \vspace{-2mm}
\end{figure}

\vspace{-1.5mm}
\spara{Agent Tool Memory.} One of our primary customers is an internal Agent platform at {\sf ByteDance}, which aims to leverage memory capabilities to enhance the reliability and efficiency of autonomous tasks. As illustrated in Figure~\ref{fig:agent_case}, the platform utilizes {\sf VikingMem} to capture historical tool usage logs and transform them into reusable operational knowledge. Since full execution traces are often verbose and noisy, the system selectively extracts each salient tool invocation as a discrete event instance, recording key information such as the tool used, the usage situation, the task goal, and its success status. These compact event instances preserve actionable signals from past executions while avoiding the redundancy of raw logs. They then evolve the persistent entity instance for that tool, which serves as a living, long-term profile rather than a static record. As new events are ingested, the entity is continuously refined to accumulate qualitative ``lessons-gleaned'' insights, including what the tool is suitable for, documented failure cases, and actionable suggestions for future invocations. This allows the agent platform to make more informed decisions, avoid repeated failures, and select the optimal tool for new tasks, thereby illustrating how {\sf VikingMem} turns raw histories into structured and evolving memory for downstream use.

\vspace{-1.7mm}
\section{Experimental Evaluation}
\label{sec:exp}
In this section, we empirically evaluate our proposed system as well as the competing approaches on effectiveness and efficiency. 
All methods are implemented in Python. The code and datasets can be found at ~\cite{code}. The vector database service was deployed on a single node with 32 vCPUs (Intel Xeon Platinum 8582C) and 8GB memory, while the embedding service ran on a single NVIDIA A30 GPU (24GB VRAM) with 16 CPU cores and 64GB memory.

\vspace{-3mm}
\subsection{Experiment Setup}
\label{subsec:setup}

\vspace{-0.3mm}
\subsubsection{Dataset}
\label{subsec:dataset}
We used two of the most widely utilized datasets for evaluating the long-term memory capabilities of large language models to assess the ability of our proposed memory system:

\noindent
\textbf{LOCOMO~\cite{maharana2024evaluating}:} a high-quality benchmark for evaluating long-term conversational memory in large language model agents. It consists of 10 conversations, each characterized by multiple dialogues with an average of 1000+ messages per conversation and a substantial number of tokens. Each conversation is annotated for the question-answering, which are categorized into different reasoning types, including single-hop, multi-hop, temporal reasoning, and open-domain knowledge.

\noindent
\textbf{LongMemEval~\cite{wu2024longmemeval}:} a longer benchmark for evaluating interactive memory in chat assistants. Following~\cite{rasmussen2025zep}, we use one of the subsets: \textit{LongMemEval\_s} for evaluation, with 500 long conversations averaging approximately 115,000 tokens in length. The token length of \textit{LongMemEval\_s} is $346\times$ larger than \textit{LOCOMO}. The dataset features timestamped conversational histories and includes hundreds of high-quality questions across different abilities like multi-session reasoning and memory updates.

\begin{table*}[!tb]
\vspace{-2mm}
\caption{ LLM-as-a-Judge evaluation scores (\%, with higher values denoting superior performance) and  Search Latency (second) for each question category in the {\em LOCOMO} and {\em LongMemEval} dataset. Scores for baselines annotated with a superscript {$\dagger$} are sourced from published evaluation results, while the search latencies for other methods in {\em LOCOMO} and {\em LongMemEval} were measured from our own experiments. Best results are in bold; second-best results are with underline. In {\em LongMemEval}, "SSU" stands for "single-session-user", "MS" denotes "multi-session", "SSP" represents "single-session-preference", "TR" indicates "temporal-reasoning", "KU" refers to "knowledge-update", and "SSA" means "single-session-assistant". } 
\vspace{-3mm}
\label{tab:performance of memory base}
\centering
\begin{tabular}{c|c|c|c|c|c|c|c|c}
\hline
\multicolumn{9}{c}{\textbf{LOCOMO}} \\ \cline{1-9}
\multirow{2}{*}{\textbf{LLM Model}} & \multirow{2}{*}{\textbf{Methods}} & \multicolumn{5}{c|}{\textbf{LLM Judge Score}} & \multicolumn{2}{c}{\textbf{Search Latency}} \\ \cline{3-9}
 {} & {} & {\textbf{Single Hop}} & {\textbf{Multi-Hop}} & {\textbf{Open Domain}} & \textbf{Temporal} & \textbf{Overall} & \textbf{p50} & \textbf{p95}
        \\ \hline \hline
\multirow{6}{*}{ \em{GPT-4o-mini} } & Mem0$^{\dagger}$ & 72.93 & 67.13 & 51.15 & 55.51 & 66.88 & \textbf{0.15} &  \textbf{0.20} \\ \cline{2-9}
                          & Mem0-graph$^{\dagger}$ & 75.71 & 65.71 & 47.19 & 58.13 & 68.44 &  0.48 & 0.66 \\ \cline{2-9}
                          & Zep$^{\dagger}$ & 74.11 & 66.04 & \underline{67.71} &  79.79 & 75.14 & 0.42 & 0.63  \\ \cline{2-9}
                          & RAG & 65.16 & 50.35 & 48.96 &  59.50 & 60.26 & 0.22 & 0.43  \\ \cline{2-9}
                          & Full-Context & 78.00 & 74.82 & 59.38 & \textbf{83.80} & 77.47 & / & / \\ \cline{2-9}
                          & Claude & 63.50 & 55.67 & 50.00 & 47.35 & 57.86 & 18.90 & 25.42 \\ \cline{2-9}
                          & Openclaw & 17.60 & 13.48 & 30.21 & 13.40 & 16.75 & 17.19 & 26.26 \\ \cline{2-9}
                          & Mirix & \underline{79.08} & \underline{76.01} & 66.67 & 80.86 &  \underline{78.66} & 10.90 &  25.76 \\ \cline{2-9}
                          & VikingMem & \textbf{94.89} & \textbf{81.91} & \textbf{78.12} & \underline{82.24} & \textbf{88.83} & \underline{0.20} & \underline{0.39}  \\
                          \hline \hline
\multirow{6}{*}{ \em{GPT-4.1-mini} } & Mem0$^{\dagger}$ & 62.41 & 57.32 & 44.79 & 66.47 & 62.47 & \textbf{0.15} & \textbf{0.20} \\ \cline{2-9}
                          & Mem0-graph & 74.44 & 68.44 & 51.04 & 54.52 & 67.73 &  0.48 & 0.66  \\ \cline{2-9}
                          & Zep$^{\dagger}$ & 79.43 & 69.16 & \underline{73.96} &  83.33 & 79.09 & 0.42 & 0.63 \\ \cline{2-9}
                          & RAG & 69.56 & 56.74 & 57.29 & 64.17 & 65.32 & 0.27 & 0.91 \\ \cline{2-9}
                          & Full-Context & 88.70 & 77.30 & 72.92 & \textbf{92.21} & \underline{86.36} &  / & /  \\ \cline{2-9}
                          & Claude & 72.29 & 68.44 & 56.25 & 55.45 & 67.08 & 10.81 & 18.62 \\ \cline{2-9}
                          & Openclaw & 26.52 & 18.09 & 26.04 & 21.81 & 23.96 & 13.73 & 19.26 \\ \cline{2-9}
                          & Mirix$^{\dagger}$ & \underline{85.11} & \underline{83.70} & 65.62 & \underline{88.39} & 85.38  &  10.90 &  25.76 \\ \cline{2-9}
                          & VikingMem & \textbf{93.46} & \textbf{85.89} & \textbf{79.79} & 88.16 & \textbf{90.12}  &  \underline{0.20} & \underline{0.39}   \\
                          \hline 
\end{tabular}
\\[0.8mm]
\begin{tabular}{c|c|c|c|c|c|c|c|c|c|c}
\hline
\multicolumn{9}{c}{\textbf{LongMemEval}} \\ \cline{1-11}
\multirow{2}{*}{\textbf{LLM Model}} & \multirow{2}{*}{\textbf{Methods}} & \multicolumn{7}{c|}{\textbf{LLM Judge Score}} & \multicolumn{2}{c}{\textbf{Search Latency}} \\ \cline{3-11}
 {} & {} & {\textbf{SSU}} & {\textbf{MS}} & {\textbf{SSP}} & \textbf{TR} & \textbf{KU} & \textbf{SSA} & \textbf{Overall} & \textbf{p50} & \textbf{p95} 
        \\ \hline \hline
\multirow{5}{*}{ \em{GPT-4o-mini} } & Mem0 & 68.57 & 41.35 & 20.00 & 37.59 & 66.67 & 5.36  & 42.80 & 0.40 &  1.00 \\ \cline{2-11}
                          & Mem0-graph & 72.86 & 38.35 & 16.67 & 34.59 & 60.26 &  35.71 & 44.00 &  1.10 & 2.09 \\ \cline{2-11}
                          & Zep$^{\dagger}$ & \textbf{92.90} & \underline{47.40} & \textbf{53.30} & \underline{54.10} & \underline{74.40} & 75.00 & \underline{63.21} & 3.80 & 5.45  \\ \cline{2-11}
                          & Full-Context & 81.43 & 40.60 & 30.00 & 36.84 & \textbf{76.92} & \underline{82.14} & 55.00 & / & / \\ \cline{2-11}
                          & VikingMem & \underline{90.00} & \textbf{53.31} & \underline{46.67} & \textbf{55.89} & 71.23 & \textbf{96.43} &  \textbf{66.36} &  0.25 & 0.89  \\
                          \hline \hline
\multirow{5}{*}{ \em{GPT-4o} } & Mem0 & 80.00 & 50.38 & 26.67 & 42.11 & 74.36 & 10.71  & 50.20 & 0.40 &  1.00 \\ \cline{2-11}
                          & Mem0-graph & \underline{84.29} & 48.87 & \underline{30.00} & 41.35 & 75.64 & 48.21 & 54.80 & 1.10 & 2.09 \\ \cline{2-11}
                          & Zep$^{\dagger}$ & \textbf{92.90} & \underline{57.90} & \textbf{56.70} & \underline{62.40} & \textbf{83.30} & 80.40 & \underline{70.40} & 3.80 & 5.45  \\ \cline{2-11}
                          & Full-Context & 81.43 & 44.36 & 20.00 & 45.11 & \underline{78.21} & \underline{94.64} & 59.20 & / & / \\ \cline{2-11}
                          & VikingMem & \textbf{92.90} & \textbf{69.92} & \textbf{56.70} & \textbf{66.92} & 76.92 & \textbf{98.21} &  \textbf{75.80} &  0.25 & 0.89  \\
                          \hline 
\end{tabular}
\vspace{-2mm}
\end{table*}

\vspace{-1.8mm}
\subsubsection{LLM Settings}
\label{subsec:llmset}
Following ~\cite{wang2025mirix, rasmussen2025zep}. We evaluated the {\em LOCOMO} dataset using \texttt{GPT-4.1-mini} and \texttt{GPT-4o-mini}, and the {\em LongMemEval} dataset using \texttt{GPT-4o-mini} and \texttt{GPT-4o}. In all configurations, the specified model served as both the answer generator and the LLM-as-a-Judge evaluator.

\vspace{-1.8mm}
\subsubsection{Baselines}
\label{subsec:baseline}

We compare \textsf{VikingMem} with eight representative long-term memory baselines covering vector retrieval, graph memory, modular memory, Markdown-first memory, native compacted memory, and brute-force full-context prompting.

\noindent
\textbf{Mem0~\cite{chhikara2025mem0}:} Extracts candidate facts from dialogue, updates memory via ADD/UPDATE/DELETE/NOOP operations, and retrieves memories with semantic search.

\noindent
\textbf{Mem0-graph~\cite{chhikara2025mem0}:} Extends Mem0 with a knowledge graph over extracted entities and relations, enabling graph-based retrieval.

\noindent
\textbf{Zep~\cite{rasmussen2025zep}:} Builds episodic, semantic, and community memory layers with temporal signals and performs hybrid retrieval with reranking.

\noindent \textbf{RAG~\cite{gao2024rag}:} This serves as a standard RAG baseline. For our comparison, dialogue history is segmented by grouping 8 messages from the same session into a single text chunk. This chunking strategy ensures that the token count of each memory unit is comparable to the granular memories managed by the other methods.

\noindent
\textbf{Mirix~\cite{wang2025mirix}:} Uses six specialized memory modules and performs hybrid retrieval across memory types.

\noindent
\textbf{Full-Context:} A naive \emph{inject-everything} baseline that directly feeds the entire available dialogue history into the model for each query, without any memory construction or retrieval.

\noindent
\textbf{OpenClaw~\cite{openclaw}:} A Markdown-first memory baseline that incrementally maintains persistent memories as text files and retrieves relevant entries at inference time.

\noindent
\textbf{Claude Native Memory:} A baseline built on Claude Code's native memory mechanism, where interaction history is periodically compacted into concise memory notes for later reuse, without explicit schema-driven retrieval.

\vspace{-2mm}
\subsubsection{Evaluation Metrics}
\label{subsec:metrics}
For the evaluation of our {\sf VikingMem}, following~\cite{wang2025mirix,chhikara2025mem0,rasmussen2025zep}, we adopt three widely used metrics: LLM-as-a-Judge (LLM Judge Score), F1-score and search latency (seconds), which are employed to evaluate the efficiency and effectiveness of our method. 
To mitigate randomness,
we run each experiment three times and report the average result for the final results.

\vspace{-2mm}
\subsubsection{Implementation Details}
\label{subsec:implement}
To reflect the latest performance of each method, we used API keys to call the memory interface of each baseline to complete our evaluation (except for the {\sf RAG} method). Since the memory ingestion cost is high, we only imported the data once per system. For each query, we repeated the answer generation and evaluation three times and reported the average results. Furthermore, to assess the practical throughput of each system, we enforced a 24-hour time limit for the entire end-to-end process of memory extraction and answering. Consequently, baselines that failed to complete the task within this timeframe were considered to have timed out, and their results are not reported. Additionally, since both datasets consist primarily of fact-based queries, where time-weighting offers little benefit, we disabled this feature by default during our experiments. For LLM-based answer generation and evaluation, all compared methods were tested under the same prompt setup. Our prompts are consistent with the evaluation protocol used in prior work~\cite{rasmussen2025zep}, ensuring a fair comparison. Meanwhile, for {\sf VikingMem}, all results were obtained on the production implementation backed by {\sf VikingDB}: event memories, entity states, and auxiliary keyword-linked indices were stored in {\sf VikingDB}, while the reported retrieval latency includes candidate generation from {\sf VikingDB} together with {\sf VikingMem}-side score fusion and reranking.

\vspace{-2mm}
\subsection{End-to-end Evaluation}
\label{sec:end-to-end_eval}
We conduct a comprehensive end-to-end evaluation of {\sf VikingMem} against state-of-the-art baselines to assess two critical dimensions of a memory system: its ability to provide accurate, relevant information (effectiveness) and its speed in real-time scenarios (efficiency). Using the public {\em LOCOMO} and {\em LongMemEval} benchmarks, we measure performance through LLM-Judge scores for effectiveness and p50, p95 search latency for efficiency, providing a holistic view of each system's practical utility. These two dimensions directly align with our design objectives from \S~\ref{subsec:design_objectives}. For each benchmark, we follow the judge settings used in prior work~\cite{rasmussen2025zep,chhikara2025mem0,li2025memos}, so as to preserve comparability with existing evaluations. Since LLM-Judge scores depend on the evaluation setup, we report them separately for each benchmark and judge model. We include two judge models to assess whether the relative ranking trend is stable across reasonable evaluation settings. In contrast, the reported efficiency metrics are measured directly from the memory system and are independent of the LLM judge. 

\vspace{-3mm}
\subsubsection{Effectiveness}
\label{sec:effectivity}
As shown in Table \ref{tab:performance of memory base}, {\sf VikingMem} consistently achieves the best overall results on both benchmarks under their respective evaluation settings, validating its effectiveness in long-term memory tasks. Our key observations are as follows:
(1) {\sf VikingMem} achieves the highest overall LLM-Judge scores across all baselines. On {\em LOCOMO}, it scores 88.83\% and 90.12\% with GPT-4o-mini and GPT-4.1-mini respectively
This suggests {\sf VikingMem} excels at not just storing facts but organizing them for effective cross-session reasoning. Its lead is maintained on {\em LongMemEval}, where it achieves top overall scores (66.36\% with GPT-4o-mini and 75.80\% with GPT-4o). A similar ranking trend is also observed under the GPT-4o setting, where {\sf VikingMem} significantly outperforms the second-best system, suggesting that while the absolute score scale varies with the judge model, the relative advantage of {\sf VikingMem} remains stable in our reported results.
(2) {\sf Mirix}, with its six specialized memory modules, is a strong second-best performer, achieving an overall score of 85.38\% on {\em LOCOMO}. However, its complex, multi-module architecture, while comprehensive, appears less streamlined than {\sf VikingMem}'s more integrated structure, resulting in a consistent performance gap.
(3) Among the stronger baselines, {\sf Zep} and {\sf Full-Context} each show competitive performance on some categories, but neither matches {\sf VikingMem} consistently. {\sf Zep}'s layered memory design is particularly effective on several {\em LongMemEval} categories, while {\sf Full-Context} remains a strong brute-force baseline on {\em LOCOMO}, confirming that directly injecting all history can be effective when the context still fits within the model's usable window. However, both fall short of {\sf VikingMem} in overall accuracy, suggesting that explicit memory organization and retrieval are more robust than either generic graph memory or raw context stuffing. This advantage is especially clear for queries requiring precise factual details and cross-turn evidence, where filtering less relevant long-context information helps the system surface the most useful evidence more reliably. Meanwhile, this trade-off remains practically important even as context windows increase: current LLMs are still susceptible to interference from irrelevant context in long inputs, which can hurt end-to-end accuracy. Even if future models become strong enough to fully overcome this issue, naively injecting the full history would still incur substantially higher token cost in production settings.
(4) Other baselines, including {\sf Mem0}, {\sf Mem0-graph}, {\sf RAG}, {\sf Claude Code Native Memory}, and {\sf OpenClaw}, perform substantially worse on most settings. {\sf Mem0}/{\sf Mem0-graph} rely on relatively brief memory units and struggle with complex multi-hop or temporal reasoning. {\sf RAG} retrieves raw conversational chunks, which often introduce substantial contextual noise. {\sf OpenClaw} performs the worst overall, likely because its Markdown-based memory is organized as coarse-grained dialogue records rather than compact structured memories, making retrieval less precise for downstream QA. {\sf Claude's Native Memory} is stronger than {\sf OpenClaw} but still lags behind ours, likely because its memory access is still largely inference-time and agent-driven, which may not always surface all relevant memory items reliably.

\vspace{-3mm}
\subsubsection{Efficiency}
\label{sec:efficiency}
Beyond accuracy, low latency is critical for user experience. Our evaluation, focusing on the order of magnitude of retrieval delays, confirms {\sf VikingMem}'s practical advantages. With a p50 latency around 0.20-0.25s and a p95 latency also remaining low at (e.g., 0.89s for {\em LongMemEval}\,), {\sf VikingMem} is not only highly responsive on average but also maintains stability under load. It operates on the same low-latency scale as lightweight methods like {\sf Mem0} and {\sf Zep}, demonstrating that its sophisticated capabilities do not compromise operational speed. In stark contrast, other high-accuracy systems make significant efficiency trade-offs. 
For instance, {\sf Mirix} exhibits p50 latencies (10.90s) that are orders of magnitude higher, with their p95 latencies climbing even further to (e.g., 25.76s), rendering it impractical for interactive applications. Furthermore, this method failed to complete the {\em LongMemEval} evaluation within the 24-hour time limit.

\vspace{-2mm}
\subsection{Effect of One-pass Extraction and \textsf{EUA}}
\label{subsec:exp_extraction}

\vspace{-0.5mm}
To quantify the benefits of our one-pass memory extraction paradigm and \textsf{EUA} algorithm described in \S~\ref{subsec:memory_ext}, we conduct a comparative analysis on the {\em LOCOMO} dataset, for which we configured three distinct memory types: one event memory and two entity memories (a user profile and a topic-based compressed memory). We benchmark our schema-driven method against the conventional multi-pass baseline, a paradigm representative of prior systems \cite{Memobase25,wang2025mirix} where each memory type is extracted in a separate LLM inference call.

We evaluate both approaches on two critical dimensions: efficiency and effectiveness. We use Extraction Cost (\$) to measure the efficiency by calculating the total monetary cost required to process and extract all defined memories from all dialogue sessions as well as the Extract Time (s). Using LLM Judge Score to assess effectiveness of the extract memories. Note that the multi-pass baseline is implemented with parallel LLM calls across memory types for a fair latency comparison.

As shown in Table~\ref{tab:one-pass}, the results validate both our one-pass extraction paradigm and the proposed \textsf{EUA} algorithm. 
First, compared with the conventional multi-pass baseline (\emph{Multiple Prompts}), one-pass extraction substantially improves cost efficiency while preserving memory quality: the extraction cost drops from \$0.35 to \$0.11 (one-pass w/o \textsf{EUA}), i.e., a 3.2$\times$ reduction, with an almost unchanged LLM-judge score.
Second, \textsf{EUA} further reduces the end-to-end latency and cost of one-pass extraction, decreasing the wall-clock time from 12.32\,s to 7.42\,s and the cost from \$0.11 to \$0.07, while maintaining comparable quality.

\begin{table}[t]
\caption{Effect of One-pass Extraction and \textsf{EUA}}
\label{tab:one-pass}
\vspace{-3.5mm}
    \centering
        \renewcommand{\arraystretch}{1.4}
       \begin{tabular}{m{2.9cm}<{\centering}|m{1.2cm}<{\centering}|m{1.2cm}<{\centering}|m{1.1cm}<{\centering}}
        \hline 
        $\textbf{Techniques}$ & $\textbf{Cost(\$)}$ & $\textbf{Time(s)}$ & $\textbf{Score}$  \\
        \hline \hline
        \em{Multiple Prompts}  & 0.35 & 11.02 & 88.86 \\
        \hline
        \em{One-pass (w/ \textsf{EUA})}  & 0.07 & 7.42 & 88.77 \\
        \hline
        \em{One-pass (w/o \textsf{EUA})}  & 0.11 & 12.32 & 88.83 \\
        \hline
      \end{tabular}
      \vspace{-3mm}
\end{table}

\vspace{-2mm}
\subsection{Storage Efficiency and Retention Analysis}
\label{subsec:exp_storage}

To evaluate retention efficiency under sparse information density, we benchmark \textsf{VikingMem} against the Naive \textsf{RAG} baseline~\cite{gao2024rag} on the \textit{LongMemEval} dataset. Performance is evaluated via two metrics: (1) \textbf{Storage Consumption}, measured by the total token count required to persist the memory state, and (2) \textbf{Retrieval Accuracy}, assessed via an LLM-Judge score to ensure that selective extraction does not compromise query correctness.




The results in Table~\ref{tab:storage_compression} demonstrate the efficacy of our selective retention strategy.

First, \textsf{VikingMem} achieves a dramatic reduction in storage costs. On the \emph{LongMemEval} dataset, the Naive RAG baseline requires storing $100\%$ of the raw tokens. In comparison, \textsf{VikingMem} retains only the extracted events and entity snapshots, reducing the total storage footprint to just 16.82\% of the original size. This corresponds to a compression ratio of approx. \textbf{6:1} (83.18\% reduction), directly validating our claim of significant cost savings.

Second, despite this massive reduction in data volume, \textsf{VikingMem} maintains, and in some cases improves, retrieval accuracy. As shown in the results, \textsf{VikingMem} achieves an LLM-Judge Score of $75.80$, comparable to or higher than the Naive RAG baseline ($63.81$). This indicates that our segmentation engine successfully separates signal from noise, proving that ``less is more'' when the retained memory is semantically dense and well-structured.

\begin{table}[t]
\caption{Storage Efficiency on LongMemEval}
\label{tab:storage_compression}
\vspace{-3.5mm}
    \centering
    \renewcommand{\arraystretch}{1.4}
    \begin{tabular}{m{2.0cm}<{\centering}|m{3.0cm}<{\centering}|m{1.5cm}<{\centering}}
    \hline
    $\textbf{Method}$ & $\textbf{Storage (Tokens)}$ & $\textbf{Score}$ \\
    \hline \hline
    \em{Naive RAG~\cite{gao2024rag}} & 100\% (Baseline) & 63.81 \\
    \hline
    \em{\textsf{VikingMem}} & \textbf{16.82\% (83.18\% $\downarrow$)} & \textbf{75.80} \\
    \hline
    \end{tabular}
    \vspace{-4mm}
\end{table}

\vspace{-1.5mm}
\subsection{Ablation Studies}
To rigorously evaluate the individual contributions of the core components within our \textsf{VikingMem} framework, we conducted a comprehensive ablation study on the {\em LOCOMO} dataset using GPT-4o-mini. As detailed in Table~\ref{tab:ablation study}, we assessed the impact of removing each module on effectiveness (LLM Judge Score) and efficiency (p95 search latency). The results clearly demonstrate that each component contributes positively, with intelligent memory segmentation and reranking being particularly critical for performance.

\subsubsection{Impact of Rerank}
The Rerank module plays a crucial role in refining search results to enhance response accuracy. Removing it caused a 3.64-point drop in the LLM Judge Score ($88.83 \rightarrow85.19$ ), confirming its importance. Meanwhile, the late-interaction mechanism rerank provides a substantial quality boost at a modest cost of only 6.8ms in p95 search latency, proving it to be an effective and efficient component for high-precision memory retrieval.

\vspace{-2mm}
\begin{table}[!tb]
\caption{ Ablation study on the contribution of each system component to end-to-end performance and its associated impact on p95 search latency, evaluated on the LOCOMO dataset using GPT-4o-mini. ``IMSM'' stands for ``\underline{i}ntelligent \underline{m}emory \underline{s}egmentation \underline{m}ethod for event-intertwined sessions''. }
\vspace{-2mm}
\label{tab:ablation study}
\centering
\begin{tabular}{c|c|c}
\hline
{\textbf{Removed}} & \multirow{2}{*}{\textbf{LLM Judge Score}} & \multirow{2}{*}{\textbf{Search Latency}} \\ 
 {\textbf{Component}} & {} & {} 
        \\ \hline \hline
 / & 88.83 & - \\ \cline{1-3}
                           Multi-Vector Rerank & 85.19 & +6.8ms \\ \cline{1-3}
                           Entity Memory & 86.93 & $\approx0$  \\ \cline{1-3}
                           IMSM & 83.51 & $\approx0$  \\ \cline{1-3}
                           Keyword Graph & 86.92 & +25.8ms  \\ \cline{1-3}
                          \hline 
\end{tabular}
\vspace{-3mm}
\end{table}

\subsubsection{Intelligent memory segmentation method}
As shown in Table \ref{tab:ablation study}. The special design of {\sf IMSM} is the most impactful component for memory quality. Its removal triggered the most severe performance drop, causing the LLM Judge Score to plummet by 5.32 points ($88.83 \rightarrow 83.51$). This significant contribution stems from its ability to solve the core challenge of noisy and interleaved dialogues by extracting only semantically rich segments and merging related but non-contiguous content into a single, coherent memory. As this is an offline process performed during memory ingestion, it provides this substantial boost to effectiveness without incurring any search latency ($\approx0$), making it a highly efficient and foundational element of our system.

\vspace{-2mm}
\subsubsection{Entity Memory}
The entity memory enhances response accuracy by leveraging long-term, evolving entity-level context. Its removal caused a 1.9-point drop in the LLM Judge Score ($88.83 \rightarrow 86.93$), confirming its value in tailoring outputs to individual users. 


\vspace{-2mm}
\subsubsection{Keyword Graph}
The Keyword Graph provides an auxiliary recall path, designed to retrieve complementary memories that the primary hybrid search might miss. Its value is confirmed by the 1.91-point drop in the LLM Judge Score upon its removal ($88.83 \rightarrow 86.92$), an impact comparable to that of the entity memory. However, it also introduces a 25.8ms search latency, making it the most computationally expensive module.

\subsection{Operator Usage Frequency Across Real-World Application Scenarios}

In this part, we analyze the usage frequency of different operators across three representative downstream application scenarios: Education, Agent Memory, and Social Companionship. The statistics are derived from entity schema definitions collected from over 100 customer applications in these three scenarios, with each customer using 2.6 schemas on average. For each scenario, we report the percentage of schemas that invoke a given operator. This analysis provides a fine-grained view of how different operators are used in practical applications and helps characterize the operator requirements of different scenarios.

\begin{table}[t]
\caption{Operator usage frequency across application scenarios. Each percentage denotes the proportion of entity schemas in the corresponding scenario that use the operator. \texttt{LLM\_M} and \texttt{TIME\_C} denote \texttt{LLM\_MERGE} and \texttt{TIME\_COMPRESS}, respectively.}
\label{tab:operator_usage_frequency}
\vspace{-3.5mm}
    \centering
    \renewcommand{\arraystretch}{1.4}
    \begin{tabular}{m{1.7cm}<{\centering}|m{0.7cm}<{\centering}|m{0.7cm}<{\centering}|m{0.7cm}<{\centering}|m{0.85cm}<{\centering}|m{0.85cm}<{\centering}|m{0.85cm}<{\centering}}
    \hline
    $\textbf{Scenario}$ 
    & $\texttt{SUM}$ 
    & $\texttt{MAX}$ 
    & $\texttt{AVG}$ 
    & $\texttt{COUNT}$ 
    & $\texttt{LLM\_M}$ 
    & $\texttt{TIME\_C}$ \\
    \hline \hline
    Education 
    & 85\% & 90\% & 96\% & 88\% & 91\% & 26\% \\
    \hline
    Agent Memory 
    & 73\% & 96\% & 98\% & 64\% & 98\% & 42\% \\
    \hline
    Social Comp. 
    & 24\% & 16\% & 32\% & 11\% & 100\% & 72\% \\
    \hline
    \end{tabular}
    \vspace{-2.5mm}
\end{table}

As shown in Table~\ref{tab:operator_usage_frequency}, different scenarios exhibit distinct operator usage patterns. Education and Agent Memory frequently use aggregation-oriented operators, especially \texttt{MAX}, \texttt{AVG}, and \texttt{LLM\_M}, indicating that these scenarios often require summarizing and consolidating accumulated user states, historical records, or execution statistics. In contrast, Social Companionship relies less on numerical aggregation, but shows strong demand for semantic merging and temporal compression, with \texttt{LLM\_M} and \texttt{TIME\_C} used in 100\% and 72\% of the schemas, respectively. Overall, these results show that the proposed operators are widely used in real downstream applications, while different scenarios require different operator combinations, which justifies our operator-based design.

\vspace{-2mm}
\subsection{Evaluation with F1-Score}

As LLM-as-a-judge scores can be sensitive to the choice of judge model and evaluation prompt, we additionally report \emph{F1-score}~\cite{goswami2025dissecting} to provide a more robust and transparent evaluation. Specifically, for all methods, we use the same answer generation model (\texttt{gpt-4o-mini}) to produce the final answer based on the retrieved memory, and then compute the token-level F1-score against the ground-truth answer on \textit{LOCOMO}. The F1-score is defined as the harmonic mean of token-level precision and recall:
As shown in Table~\ref{tab:f1-score}, VikingMem achieves the best overall F1-score of 54.98\%, outperforming all baselines, including Full-Context (53.03\%). It also performs best on Single-Hop, Multi-Hop, and Open-Domain questions, while remaining competitive on Temporal questions. These results are consistent with the conclusions drawn from the LLM judge evaluation, further supporting the effectiveness of VikingMem.

\begin{table}[!tb]
\caption{F1-score (\%) on LOCOMO.
SH: Single-Hop, MH: Multi-Hop, OD: Open-Domain, Temp: Temporal.
}
\vspace{-3mm}
\label{tab:f1-score}
\centering
\small
\begin{tabular}{c|c|c|c|c|c}
\hline
\textbf{Methods} & \textbf{SH} & \textbf{MH} & \textbf{OD} & \textbf{Temp} & \textbf{Overall} \\ 
\hline \hline
Mem0 & 47.65 & 38.72 & 28.64 & 48.93 & 45.09 \\ \cline{1-6}
Mem0-graph & 49.27 & 38.09 & 24.32 & 51.55 & 46.14 \\ \cline{1-6}
Zep & 49.56 & 35.74 & \underline{41.37} & 52.04 & 47.03 \\ \cline{1-6}
Full-Context & \underline{55.64} & \underline{43.52} & 40.43 & \textbf{58.32} & 53.03 \\ \cline{1-6}
Claude & 41.23 & 34.23 & 28.06 & 46.32 & 40.19 \\ \cline{1-6}
Openclaw & 20.12 & 11.36 & 10.04 & 21.32 & 18.14 \\ \cline{1-6}
VikingMem & \textbf{59.59} & \textbf{44.52} & \textbf{43.13} & \underline{55.62} & \textbf{54.98} \\
\hline
\end{tabular}
\vspace{-2mm}
\end{table}

\section{Related Work}
\label{sec:related}

\subsection{Retrieval-Augmented Generation}
Retrieval-Augmented Generation (RAG) integrates external knowledge retrieval to mitigate LLM hallucinations and enhance factual accuracy without retraining~\cite{lewis2020retrieval,gao2024rag,zhao2024retrieval}. Modern variants extend this paradigm via modular query rewriting~\cite{ma2023query}, graph structures for relational precision~\cite{he2024g,huang25kdd}, agentic iterative reasoning~\cite{singh2025agentic}, and multimodal inputs~\cite{caffagni2024wiki}. These methods have shown great gains on benchmarks like HotpotQA~\cite{yang2018hotpotqa}. However, traditional RAG remains constrained in long-term memory management. Its reliance on static, chunk-based retrieval struggles with low-information-density memory streams, where critical user details are scattered across non-contiguous segments, often leading to noisy or incomplete extractions~\cite{wang2024adaptive}.

\subsection{Memory-centric Architecture and Systems}
To bypass finite context windows, recent memory-centric architectures implement specialized storage layers to enable coherent, long-term LLM interactions~\cite{packer2023memgpt,jones2024designing}. Representative systems like \textsf{Mem0}~\cite{chhikara2025mem0} and \textsf{MemOS}~\cite{li2025memos} dynamically extract salient facts or manage memory lifecycles, while \textsf{Zep}~\cite{rasmussen2025zep} and \textsf{MIRIX}~\cite{wang2025mirix} utilize hierarchical structures and specialized modules to preserve granular contextual relevance.

Despite these advances, existing designs fall short of a general-purpose, task-agnostic memory substrate. Most are strictly optimized for companion-style dialogue flows, failing to natively support the procedural, auditable traces required by autonomous agents or the fine-grained preference trajectories in education and recommendation domains~\cite{chhikara2025mem0,Memobase25}. Furthermore, heavy operations such as dynamic knowledge-graph maintenance under continuous memory influx introduce substantial build-time latency and resource overhead, conflicting with the strict cost and performance constraints of real-world production deployments~\cite{rasmussen2025zep,wang2025mirix}.

\vspace{-0.8mm}
\section{Conclusion} %
\vspace{-0.2mm}
\label{sec:concl}
In this paper, we addressed the fundamental data management challenge facing the next generation of LLM applications, which are shifting from stateless queries to long-term, stateful interactions. We proposed the Memory Base, a new general framework designed specifically to manage persistent state. This paradigm moves beyond static repositories, providing inherent capabilities for selective memory extraction, continuous stateful evolution, and a generalizable design. We presented {\sf VikingMem}, an end-to-end Memory Base Management System that operationalizes this framework. The core of {\sf VikingMem}'s design is its interconnected event and entity abstractions, which use intelligent segmentation to capture high-value, episodic events and a flexible, operator-based mechanism to progressively evolve persistent entity states. Extensive evaluations on public benchmarks confirmed that {\sf VikingMem} achieves SOTA retrieval effectiveness, low latency, and low extraction cost, validating {\sf VikingMem} as a robust, efficient, and scalable solution for enabling long-term contextual intelligence in LLM applications.




\begin{acks}
The authors are deeply grateful to colleagues from the \texttt{Viking} team for their foundational contributions and insightful discussions. This work was supported in part by the NSFC (Grants No. 62502434, U23A20296, 62502426), Zhejiang Province's ``Jianbing'' R\&D Project (Grant No. 2025C01195), and Yongjiang Talent Introduction Programme (Grant No. 2022A-237-G). Xiangyu Ke is the corresponding author of this work.
\end{acks}

\balance
\bibliographystyle{ACM-Reference-Format}
\bibliography{ref}


\end{document}